\documentclass[11pt]{article}
\PassOptionsToPackage{table}{xcolor}
\usepackage[preprint]{acl}
\usepackage{times}
\usepackage{latexsym}
\usepackage[T1]{fontenc}
\usepackage[utf8]{inputenc}
\usepackage{microtype}
\usepackage{inconsolata}
\usepackage{amsmath,amssymb,amsthm,bm}
\usepackage{graphicx,booktabs,array,tabularx}
\usepackage{placeins}
\definecolor{PaperAccent}{HTML}{2166AC}
\definecolor{PaperInk}{HTML}{20333E}
\definecolor{PaperMuted}{HTML}{6C7D89}
\definecolor{TableTint}{HTML}{EDF6F6}

\newtheorem{proposition}{Proposition}

\newcommand{\figinput}[1]{\includegraphics[width=\linewidth]{figures/#1}}

\title{Understanding MoE Decisions Beyond Gradients}
\author{
    Zhen Xu \\
    University of Chicago \\
    \And
    Zitao Zhang \\
    eBay \\
    \And
    Kevin Zhao \\
    eBay \\
    \And
    Xuefeng Liu \\
    University of Florida
}

\begin{document}
\maketitle
\begin{abstract}
Understanding a language model's decision requires knowing how its internal
computations affect its preference between answers. Gradient-based methods
estimate this influence by predicting the effect of changing those
computations. In mixture-of-experts (MoE) models, replacing one expert can
change which experts later layers select. Those switches can alter the
answer scores in ways that derivatives do not capture. We introduce a
predictor that uses the model's own expert outputs and selection rules to
estimate these extra effects, then adds them to a second-order prediction.
Across questions from three benchmarks, our method reduces the error in
predicting answer-score changes by 60\% on Qwen3.6 and 47\% on
Gemma4, relative to a multistep baseline using gradients and second
derivatives.
\end{abstract}

\section{Introduction}
Knowing a language model's answer does not tell us which computations
shaped it. One way to investigate is to change part of the model and measure
how its preference between answers changes. Such interventions provide
evidence about the role of internal computations
\citep{meng2022rome,wang2022ioi,conmy2023acdc}. Gradient-based attribution
predicts intervention effects and helps select which ones to examine
\citep{syed2024eap,kramar2024atp}. If a large effect receives a small predicted
score, an analysis may skip a test that would have revealed something
important about the model's behavior.

Mixture-of-experts (MoE) language models make this prediction difficult in
a particular way. Each routed layer selects a few expert networks for the
current token \citep{shazeer2017moe}. Replacing one selected expert changes
the input to later layers, which may then select different experts.
The final effect depends on both \emph{how expert outputs change} and
\emph{which experts the model uses afterward}. Even when router scores
change smoothly, swapping two different expert functions can make the
answer score jump. Local derivatives need not capture that jump.

For a question with candidates A and B, this gives a concrete test:
predict how replacing an expert shifts their score gap, including the
consequences of any later expert switches.

Second-order estimates and additional gradient samples can describe a
smooth response more accurately \citep{zhang2026hvp}. Their intermediate
samples can even use different experts, yet summing the derivatives can
still miss the effect of crossing from one expert set to another.
Prior work shows that fixing expert choices can reduce a gap between
integrated-gradient attributions and the measured response
\citep{nguyen2026contract}. We address the prediction problem with those
choices free to change.

Our approach keeps the information supplied by derivatives and adds an
estimate of the effect of later expert choices. We build this estimate
from the model's own expert outputs and selection rules, without fitting
to measured intervention outcomes. Across Qwen3.6, Gemma4, and three
question collections, the resulting predictor lowers error in all six
model--dataset comparisons (Table~\ref{tab:summary-mae}).
The gain also changes which interventions an analysis finds: on Qwen,
it recovers 79.2\% of the true top five expert combinations, compared with
63.3\% for the five-step second-order baseline
(Table~\ref{tab:summary-ranking}). In actual model runs, keeping later
expert choices fixed makes the baseline nearly exact while retaining
over 90\% of the mean intervention magnitude.

\section{Predicting the effect of an expert intervention}\label{sec:short-method}
\paragraph{Following the model's preference.}
For a prompt $x$, we track the score of a reference answer token $a$
relative to a competing output $b$. If $\ell$ denotes the model's logits,
the intervention effect is
\begin{equation}
 \begin{aligned}
 F_x(I)&=\ell_a(x;I)-\ell_b(x;I),\\
 \Delta(I)&=F_x(I)-F_x(0).
 \end{aligned}
 \label{eq:short-target}
\end{equation}
A positive $\Delta$ shifts the scores toward $a$ relative to $b$; a negative
one shifts them toward $b$. The competitor is the highest-scoring
alternative among the option tokens and retained high-scoring nonoption
tokens in the original run. Both tokens stay fixed across interventions.

At the final prompt position, $I$ replaces one active expert at each of
one, two, or three designated layers. The remaining designated layers
keep their original expert sets; other layers select experts normally.
The model uses its original normalization rules. We predict the resulting
score change, whether or not the original answer is correct.

\paragraph{The full predictor.}
We interpolate between the original and replacement outputs at the edited
layers. HVP1 uses the gradient and second derivative at the original end
of that path \citep{zhang2026hvp}. We then estimate the same intervention
in a response model twice. In $S_{\rm nat}$, later layers can choose
different experts. In $S_{\rm fix}$, they keep the same experts, but their
weights can change. Our prediction is
\begin{equation}
 \widehat\Delta(I)=\widehat\Delta_{\rm HVP1}(I)
                  +S_{\rm nat}(I)-S_{\rm fix}(I).
 \label{eq:short-ours}
\end{equation}
The difference estimates the extra effect of allowing later expert choices
to change. Both terms come from the response model below. We use the same
formula for every question, with no fitted correction strength.

\paragraph{Following later expert choices.}
From the unmodified run, we record router scores and expert outputs,
including candidate experts that were not selected. We also compute
derivatives linking expert contributions to later router scores and the
final score difference. These quantities define a response model that can
be reused for different interventions on the same prompt.

Let $z_s^0$ be the original scores at router $s$, $g_s$ its gate, and
$L_s=Dg_s(z_s^0)$ its local derivative. The predicted change $d_s$ in router
scores gives a gate response beyond the local linear estimate:
\begin{equation}
 \begin{aligned}
 r_s&=g_s^{I}(z_s^0+d_s)-g_s(z_s^0)-L_s d_s,\\
 d_t&=\sum_{s<t}H_{ts}r_s,\qquad S(I)=\sum_s R_s r_s.
 \end{aligned}
 \label{eq:short-recurrence}
\end{equation}
Here $H_{ts}$ and $R_s$ combine recorded expert outputs with derivatives
linking layer $s$ to router $t$ and to $F_x$. The gate returns sparse expert
weights. It applies the assigned sets at the three intervention layers and
the chosen natural or fixed rule elsewhere. Subtracting $L_s d_s$ avoids
counting the local gate response twice.

A forced replacement first changes $r_s$. This changes later router scores,
which can bring another expert into use and create a further response.
The recurrence follows this sequence in layer order. Expert outputs and
the intervening computation are approximated around the original run;
Gemma's mixture normalization is included
(Appendix~\ref{sec:method}).

\paragraph{Why extra derivatives are not enough.}
A smooth change in router scores can cross a point where one expert
replaces another. The answer score can jump at that point. Summing
derivatives between such points need not recover the jumps, even when
routing is recomputed at every sample. Appendix~\ref{sec:problem} gives
the formal decomposition. Our added response estimates the effect of
allowing expert choices to change, including what happens after a switch.

\begin{table*}[t]
\centering
\begingroup
\footnotesize
\renewcommand{\arraystretch}{1.10}
\setlength{\tabcolsep}{3pt}
\begin{tabular*}{\linewidth}{@{\extracolsep{\fill}}llrrrrrr@{}}
\toprule
Model & Dataset & AtP & HVP1 & HVP5 & Ours & Reduction & \shortstack{HVP5 $-$ ours\\(mean $\pm$ SE)} \\
\midrule
\textbf{Qwen3.6} & MMLU-Pro & 0.0676 & 0.0629 & 0.0626 & \textbf{0.0250} & 60.0\% & $0.0375 \pm 0.0084$ \\
 & SuperGPQA & 0.0557 & 0.0521 & 0.0500 & \textbf{0.0185} & 63.0\% & $0.0315 \pm 0.0055$ \\
 & HLE & 0.0357 & 0.0334 & 0.0327 & \textbf{0.0125} & 61.7\% & $0.0202 \pm 0.0029$ \\
\midrule
\textbf{Gemma4} & MMLU-Pro & 0.1478 & 0.1209 & 0.1144 & \textbf{0.0602} & 47.4\% & $0.0542 \pm 0.0111$ \\
 & SuperGPQA & 0.0975 & 0.0790 & 0.0722 & \textbf{0.0412} & 43.0\% & $0.0310 \pm 0.0053$ \\
 & HLE & 0.1031 & 0.0836 & 0.0653 & \textbf{0.0507} & 22.4\% & $0.0146 \pm 0.0094$ \\
\bottomrule
\end{tabular*}
\endgroup

\caption{\textbf{Predicting changes in answer preference.} MAE in logit
units; lower is better. All methods use the same 13 interventions per question.
Reduction is relative to HVP5; the last column is HVP5 MAE minus ours.
Full comparisons and per-question observations are in
Appendix~\ref{sec:results}.}
\label{tab:summary-mae}
\end{table*}

\section{Experimental setup}\label{sec:short-setup}
We test whether the predictor finds influential interventions and estimates
their effects accurately. Qwen3.6-35B-A3B and Gemma4-26B-A4B-it
\citep{qwen2026,gemma2026} receive the same 158 questions: 56 from MMLU-Pro,
52 from SuperGPQA, and 50 text-only multiple-choice questions from HLE
\citep{wang2024mmlu,du2025supergpqa,phan2025hle}. These subsets cover multiple
subjects and were fixed without filtering on model performance.
Both models select eight experts per routed layer, from pools of 256 and
128 experts. Their parameters stay fixed throughout.

\paragraph{Selecting important interventions.}
On 24 of these questions (eight per dataset), each method chooses five of
32 expert combinations: 16 pairs and 16 triples. We then check those choices
against effects measured by running the edited model. Recall@5 is the
fraction of the true five largest absolute effects that the method recovers.
Effect-mass capture is the total absolute effect of the selected five,
relative to the best possible five.

\paragraph{Estimating effect sizes.}
Across all 158 questions, we compare predictions on 13 matched interventions:
nine single replacements and four combinations. Replacements span weakly
to strongly weighted active experts and high-scoring inactive experts.
MAE averages prediction errors first within each question and then equally
across questions. Thus MAE tests numerical accuracy, while ranking tests
which interventions would reach an analysis.

\paragraph{Comparisons and controls.}
We adapt first-order AtP, second-order HVP1, and five-step HVP5 to the same
replacement path. HVP5 recomputes routing at each sampled position.
The single-sum reference separately measures single-replacement effects and
adds them to predict combinations. Further comparators and implementation
details are in Appendix~\ref{sec:evaluation}.

On the 24-question panel, we also run the actual model with later expert
identities fixed, allowing their weights and all other computation to
respond. HVP20 uses twenty derivative samples on eight prespecified joint
interventions per question. Reported gains are paired means $\pm$ SE,
stratified by subject or dataset. Sampling and execution details are in
Appendices~\ref{sec:evaluation} and~\ref{app:numerical}.

\section{Results and analysis}\label{sec:short-results}
\begin{table}[t]
\centering
\begingroup
\footnotesize
\renewcommand{\arraystretch}{1.10}
\setlength{\tabcolsep}{1.5pt}
\begin{tabular*}{\linewidth}{@{\extracolsep{\fill}}lrrrr@{}}
\toprule
 & \multicolumn{2}{c}{Qwen3.6} & \multicolumn{2}{c}{Gemma4} \\
\cmidrule(lr){2-3}\cmidrule(l){4-5}
Predictor & Recall & Mass & Recall & Mass \\
\midrule
\multicolumn{5}{l}{\emph{Predictors without measured single effects}} \\
AtP & 54.2 & 79.3 & 62.5 & 85.3 \\
HVP5 & 63.3 & 84.3 & 78.3 & 93.5 \\
\textbf{Ours} & \textbf{79.2} & \textbf{93.1} & \textbf{83.3} & \textbf{97.5} \\
\addlinespace[2pt]
Ours $-$ HVP5 & $15.8 \pm 6.5$ & $8.8 \pm 4.4$ & $5.0 \pm 4.5$ & $4.0 \pm 2.1$ \\
\midrule
\multicolumn{5}{l}{\emph{Reference using measured single effects}} \\
Single sum & 69.2 & 90.0 & 76.7 & 94.3 \\
Ours $-$ sum & $10.0 \pm 5.2$ & $3.1 \pm 2.7$ & $6.7 \pm 2.8$ & $3.2 \pm 1.3$ \\
\bottomrule
\end{tabular*}
\endgroup

\caption{\textbf{Selecting combinations with the largest effects on answer
scores.} Recall@5 and captured effect mass (\%) among 32 joint candidates.
Difference rows are Ours minus the named comparator, reported as paired
mean $\pm$ SE in percentage points, calculated before rounding.
Single sum adds measured single-replacement effects; the predictors above
do not use these measurements.}
\label{tab:summary-ranking}
\end{table}

\begin{figure*}[t]
\centering\figinput{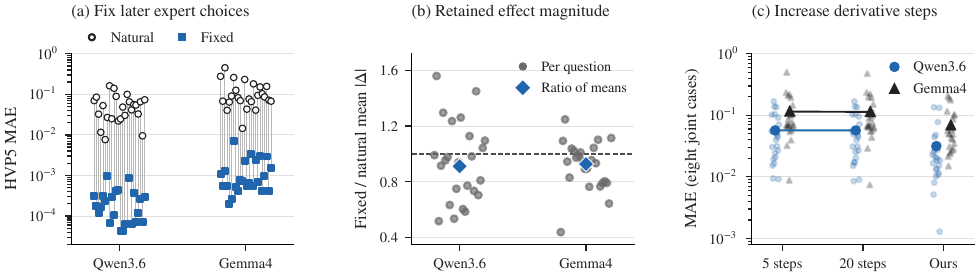}
\caption{\textbf{Later expert choices are a source of prediction error.}
(a) Keeping the same experts selected sharply reduces HVP5 error; segments pair
the same question. (b) Most mean effect magnitude remains. Dots show
question-level ratios; diamonds show ratios of model means.
(c) Increasing HVP from five to twenty steps barely helps on the same eight
joint cases. Small points are questions, large markers are means.
Each model uses 24 questions; (a,b) use all 32 joint candidates.}
\label{fig:short-mechanism}
\end{figure*}

\paragraph{Better predictions reveal more influential interventions.}
When an analyst chooses interventions by their predicted effects, ranking
errors determine which changes get examined. On Qwen, our selected set
contains 3.96 of the true top five combinations on average, compared with
3.17 for HVP5 (Table~\ref{tab:summary-ranking}). Recall rises from 63.3\% to
79.2\%, a paired gain of $15.8\pm6.5$ percentage points. Gemma's recall
rises from 78.3\% to 83.3\%, with a paired gain of $5.0\pm4.5$ points.
The improved estimates therefore help an analysis recover important
interventions it would otherwise leave out.

The selected interventions also capture more of the available effect
magnitude: 93.1\% versus 84.3\% on Qwen, and 97.5\% versus 93.5\% on Gemma.
This measure gives credit for selecting a near-equally influential
alternative to a true top-five intervention. The separate single-sum
reference uses measured effects and reaches recall of 69.2\% and 76.7\%.
Our paired gains over this reference are $10.0\pm5.2$ and $6.7\pm2.8$ points.
Each measured single effect already includes its
own later routing changes, but adding those effects does not fully predict
what happens when replacements are combined.

\paragraph{The accuracy gain extends across all six settings.}
Table~\ref{tab:summary-mae} compares effect predictions on identical
questions and interventions. Relative to HVP5, our method reduces MAE by
60--63\% across Qwen's three collections and 22--47\% across Gemma's.
The smallest advantage is on Gemma--HLE:
$0.0146\pm0.0094$ logit units, with lower error on 36 of 50 questions.
The complete comparisons and individual question errors are in
Appendix~\ref{sec:results}.

\paragraph{Holding later choices fixed removes most of the error.}
We test the role of expert switching by running each intervention with
later expert identities fixed to their original sets. Their weights and
all other computation remain free to respond. This is a separate test in
the actual model of the mechanism our predictor tries to capture.

HVP5 error falls from 0.0553 to 0.000249 on Qwen and from 0.1215 to
0.001385 on Gemma, decreasing on every question
(Figure~\ref{fig:short-mechanism}a). Meanwhile, the mean absolute score
effects retain 91.2\% and 92.8\% of their natural-routing magnitudes
(panel b). The scores still change by almost as much on average, while most
of the prediction error disappears. This connects the baseline's error
to changes in expert selection.

\paragraph{More derivative samples leave the gap.}
On eight matched joint interventions per question, increasing from HVP5
to HVP20 changes MAE from 0.05692 to 0.05691 on Qwen and from 0.11515
to 0.11344 on Gemma (Figure~\ref{fig:short-mechanism}c). Our errors on those
same interventions are 0.03156 and 0.06907. The paired advantages over
HVP20 are $0.0254\pm0.0106$ and $0.0444\pm0.0133$.
Both derivative baselines recompute routing at their sample points.
The additional samples still leave the gap that our estimate of expert
switching helps close.

\paragraph{Both parts of the predictor help.}
The full response model alone gives MMLU-Pro errors of 0.0317 on Qwen and
0.0873 on Gemma. Combining its natural-versus-fixed difference with HVP1
reduces these to 0.0250 and 0.0602, and improves all six model--dataset
means. Holding expert choices fixed inside the response model instead
gives errors close to AtP. These comparisons support keeping both the local
derivative estimate and the extra response from expert switching
(Appendices~\ref{sec:results} and~\ref{sec:ranking}).

\paragraph{Computation.}
Building the response model and evaluating HVP1 for each intervention
currently cost more than direct edited forward passes in our
cached-suffix implementation (Appendix~\ref{app:reproduction}).
Reducing this cost is an important next step toward practical use.

\section{Conclusion}
The effect of replacing an expert depends on both its output and the
experts the model selects afterward. Estimating those later choices makes
intervention predictions more accurate and helps identify influential
expert combinations. This provides better evidence about which changes
affect a model's preference between answers. For MoE models, understanding
expert influence benefits from tracking changes in the computation itself
alongside the information supplied by gradients.

\newpage
\section*{Limitations}

Our experiments cover two MoE models, three multiple-choice
question subsets, and expert replacements at three selected
layers of the final prompt position. We predict changes in
a fixed next-token logit contrast, rather than answer
correctness or generated behavior. The structural response
also relies on clean-input expert outputs and local
derivatives, which may be less accurate for large
interventions.

The method requires white-box access and per-intervention
curvature computation. In our current implementation, it
is more expensive than directly executing the tested
interventions. Finally, the main results use FP32, while
earlier BF16 measurements show smaller gains.
Generalization to lower-precision settings requires
further study.
\label{main:end}
\label{references:start}
\bibliography{references}
\appendix
\section{What is being predicted?}\label{sec:problem}
\subsection{Expert replacement at fixed source layers}
Fix a prompt $x$ and the next-token computation at its final position.
At routed layer $t$, a router assigns scores $z_t$ to candidate experts and
selects a set of $k$ experts. A sparse gate vector $g_t(z_t)$ specifies their
weights. We intervene on the selected expert identities at three designated
source layers. Each edited source replaces one of its clean active experts
with a previously inactive expert. The model recomputes the weights for the
assigned set using its native normalization rule.

An intervention $I=(a_1,a_2,a_3)$ specifies one menu entry per source.
The zero entry assigns the clean expert set. All three sources are explicitly
assigned in every condition, including sources with zero entries. This fixes
the background of a single replacement and of a pair or triple consistently.
At other layers, experts are selected normally unless an experiment explicitly
fixes those selections. Shared experts and the remaining computation retain
their native implementation.

The unit of prediction is a \emph{prompt--intervention pair}. We do not learn
a single new routing policy across prompts, and the intervention is not a
complete replacement of all routing decisions. Its downstream consequences
are part of the effect to be predicted.

\subsection{A fixed next-token logit contrast}
For each prompt, let $a$ be the dataset's reference answer token and let $b$ be
the highest-scoring nonreference token in the retained clean readout set.
The readout set contains the option tokens and high-scoring clean nonoption
tokens. Both $a$ and $b$ are fixed before intervention. Define
\begin{equation}
 \begin{aligned}
 F_x(I)&=\ell_a(x;I)-\ell_b(x;I),\\
 \Delta_x(I)&=F_x(I)-F_x(0),
 \end{aligned}
 \label{eq:target}
\end{equation}
where $\ell$ is the native model's next-token logit vector. A predictor returns
a real number $\widehat\Delta_x(I)$ in logit units. Its absolute error is
$|\widehat\Delta_x(I)-\Delta_x(I)|$.

Fixing the two tokens prevents a moving maximum over rivals from introducing
another source of nonsmoothness. The choice of score and intervention is
part of the explanation being tested \citep{zhang2024best,heimersheim2024patching}. This score can be measured on a model that
answers the question incorrectly: correctness is not required to define the
intervention effect. The reference label specifies a diagnostic readout. Our
evaluation measures prediction of that readout, rather than benchmark answer
accuracy, generation quality, or success at forcing a target answer.

\subsection{Derivative predictions along a native intervention path}
We embed a discrete replacement in a continuous path. At an active source,
evaluate its clean-assigned and edited-assigned expert branches on the current
input, then interpolate their outputs with parameter $\lambda$.
The endpoints $\lambda=0$ and $\lambda=1$ implement the clean and edited
assignments. In Gemma, interpolation occurs after the native post-mixture
normalization. All active sources move together along
$\boldsymbol\lambda=t\mathbf{1}$, giving a scalar path $f_I(t)$.
The endpoint relation $f_I(1)-f_I(0)=\Delta_x(I)$ is checked against native
execution, including the complete output vocabulary.

In these coordinates, first-order attribution prediction and its one-step
second-order extension are
\begin{equation}
 \begin{aligned}
 \widehat\Delta_{\mathrm{AtP}}&=f_I'(0),\\
 \widehat\Delta_{\mathrm{HVP1}}&=f_I'(0)+\tfrac12 f_I''(0).
 \end{aligned}
 \label{eq:hvp}
\end{equation}
For a joint edit, $f_I''(0)=\mathbf{1}^{\top}H_I(0)\mathbf{1}$ retains
mixed-source Hessian terms. Multistep HVP uses
\begin{equation}
 \begin{aligned}
 \widehat\Delta_{\mathrm{HVP}K}
 =\sum_{j=0}^{K-1}\Bigl[
 &\frac{f_I'(j/K)}{K}\\[-2pt]
 &+\frac{f_I''(j/K)}{2K^2}\Bigr].
 \end{aligned}
 \label{eq:multistep}
\end{equation}
This is the multistep formula of \citet{zhang2026hvp}, adapted to the declared
expert-interpolation path. The native routers are recomputed at every sampled
intermediate state. HVP5 therefore does not hold routing fixed along the path.

\subsection{Why more derivative samples can leave an error}
Consider the real-arithmetic idealization of this path. Within an interval
with fixed expert identities, $f_I$ is smooth. At a top-$k$ boundary, replacing
one expert function by another can produce a finite jump in $f_I$.
The following elementary decomposition explains the distinction.

\begin{proposition}[Smooth response and selection jumps]\label{prop:jumps}
Suppose $f$ is piecewise $C^2$ on $[0,1]$ with finitely many internal
boundaries $\tau_1,\ldots,\tau_J$, finite one-sided values, and bounded
first and second derivatives on the smooth pieces. There are no endpoint
jumps. Then
\begin{equation}
 \begin{aligned}
 f(1)-f(0)={}&\int_0^1 f'(t)\,dt\\
 &+\sum_{j=1}^{J}\bigl[f(\tau_j^+)-f(\tau_j^-)\bigr].
 \end{aligned}
 \label{eq:jumpdecomposition}
\end{equation}
As the uniform grid is refined, the expression in
Equation~\eqref{eq:multistep}, using bounded one-sided derivatives where
needed, converges to the integral term. A nonzero net jump therefore remains
outside that derivative approximation.
\end{proposition}
\begin{proof}
Integrate $f'$ on each smooth interval and sum the endpoint differences.
The internal one-sided values telescope after adding the displayed jumps.
The first-derivative Riemann sum converges to the integral; the magnitude
of the summed second-derivative correction is bounded by
$\sup|f''|/(2K)$ and tends to zero.
\end{proof}

For a globally $C^3$ path with $|f'''|\le M$, the usual Taylor remainder
instead gives an HVP$K$ error bound $M/(6K^2)$. Sparse selection can violate
that smoothness premise. A route change need not create a large output jump:
expert outputs or downstream effects may agree or cancel. The empirical
question is how much this distinction matters for the interventions under
study. The fixed-selection and denser-sampling experiments in
Appendix~\ref{sec:mechanism} directly test it.

\section{A structural correction for downstream routing}\label{sec:method}
The correction estimates what changes when later layers are allowed to
choose different experts. For the same source intervention, we compare a
response with downstream expert reselection against one that keeps the clean
expert sets. Their difference supplies the extra response to add to a local
second-order estimate. Both responses come from the compact approximation
below; measured edited-model outputs are used only for evaluation.

\subsection{Retaining the native gate in a local response model}
We compile a prompt-specific response model from a clean native execution.
Let $z_t^0$ be the clean scores at router $t$, and let $E_t$ contain the
outputs of its candidate experts at the clean expert input, expressed at the
native expert write site. These outputs include inactive candidate experts.
Let $L_t=Dg_t(z_t^0)$ be the clean gate derivative. For a predicted router-score
change $d_t$ and assignment $A_t$, define the finite gate remainder
\begin{equation}
 \begin{aligned}
 r_t(d_t;A_t)={}&g_t^{A_t}(z_t^0+d_t)\\
 &-g_t(z_t^0)-L_t d_t.
 \end{aligned}
 \label{eq:remainder}
\end{equation}
The gate $g_t^{A_t}$ uses the declared assigned set at an intervention source;
at an unrestricted router it uses the native top-$k$ rule. The subtraction
removes the smooth gate contribution already included in the clean Jacobian.

Let $T_{ts}$ be the clean Jacobian mapping a write at expert site $s$ to scores
at later router $t$, and let $U_s$ map that write to the fixed output contrast.
Set $H_{ts}=T_{ts}E_s$ and $R_s=U_sE_s$. For the membership interventions
studied here, the compiled response satisfies
\begin{equation}
 \begin{aligned}
 d_t&=\sum_{s<t}H_{ts}r_s(d_s;A_s),\\
 S(I)&=\sum_s R_s r_s(d_s;A_s).
 \end{aligned}
 \label{eq:recurrence}
\end{equation}
The first forced source can have a nonzero remainder even when its incoming
score displacement is zero. Its write changes the predicted scores at later
routers; those routers evaluate the native gate again. The strictly causal
block structure allows one pass through the routers.

The clean Jacobians include smooth paths through attention, normalization,
active expert content, shared branches, and gate weights. The nonlinear
correction retains finite changes in the gate coordinates while evaluating
expert output bases at their clean inputs. Consequently, the recurrence is
an approximation to the native finite response, not a full forward pass.
Its coefficients can be reused for several interventions on the same prompt.

Equation~\eqref{eq:recurrence} follows by eliminating hidden residual
coordinates from a local model with clean linear transport and the additional
writes $E_s r_s$. Before router $t$, each earlier write contributes
$T_{ts}E_s r_s$; summing them gives its score displacement. This derivation
also explains why the derivative subtraction in Equation~\eqref{eq:remainder}
is necessary.

\subsection{Two responses and one correction}
We evaluate the same compiled model with two downstream selection rules:
\begin{itemize}
 \item $S_{\mathrm{natural}}(I)$ recomputes the expert sets at nonsource
 routers from the predicted scores.
 \item $S_{\mathrm{fixed}}(I)$ assigns the clean expert sets at those routers,
 while recomputing their weights from the predicted scores.
\end{itemize}
Both evaluations apply the same intervention at the designated sources.
Our complete predictor is
\begin{equation}
 \boxed{\begin{aligned}
 \widehat\Delta(I)={}&\widehat\Delta_{\mathrm{HVP1}}(I)\\
 &+S_{\mathrm{natural}}(I)-S_{\mathrm{fixed}}(I).
 \end{aligned}}
 \label{eq:ours}
\end{equation}
The HVP term models local smooth curvature. The structural difference
estimates the extra response permitted by downstream expert reselection.
The unit coefficients in Equation~\eqref{eq:ours} were fixed on earlier
development work and are unchanged across the reported cohorts. There is no
regression on measured intervention outcomes, no learned per-question
calibration, and no selection of correction strength on the evaluation data.

The structural difference is a finite natural-versus-fixed response
approximation. It includes subsequent smooth consequences of changed expert
selection and need not equal only the jump sum in
Equation~\eqref{eq:jumpdecomposition}. This distinction gives the method a
concrete target: approximate the response caused by allowing downstream
expert reselection, including its propagation.

Let $\Delta_{\mathrm{fixed}}$ denote the effect measured in a native execution
with fixed nonsource expert sets. Adding and subtracting this quantity gives
the error identity
\begin{align}
 \Delta_{\mathrm{natural}}-\widehat\Delta
 ={}&\bigl(\Delta_{\mathrm{fixed}}-
             \widehat\Delta_{\mathrm{HVP1}}\bigr) \nonumber\\
 &+(\Delta_{\mathrm{natural}}-\Delta_{\mathrm{fixed}}) \nonumber\\
 &-(S_{\mathrm{natural}}-S_{\mathrm{fixed}}).
 \label{eq:erroridentity}
\end{align}
At a clean state with stable top-$k$ sets, the natural and fixed executions
have the same local derivative branch. The first term therefore concerns
smooth-response approximation, while the second concerns prediction of the
additional finite response under natural selection. This identity motivates
the correction; its smallness is an empirical question.

\subsection{Gemma's normalized expert mixture}
The expert-coordinate map must preserve the architecture's actual write.
Gemma applies learned expert scales and a normalization after the routed
mixture. Let $B$ contain the scaled clean expert outputs as rows and let
$\beta(z)$ be the sparse normalized gate probabilities. For hidden width $d$,
$K=BB^\top/d$, and normalization weight $\theta$, the clean-input mixture
can be written
\begin{equation}
 \begin{gathered}
 F\nu(z),\qquad F=\operatorname{diag}(\theta)B^\top,\\
 \nu(z)=\frac{\beta(z)}{\sqrt{\beta(z)^\top K\beta(z)+\varepsilon}}.
 \end{gathered}
 \label{eq:gemma}
\end{equation}
We use $F$ as the write basis and $\nu$ as the gate coordinate in
Equations~\eqref{eq:remainder}--\eqref{eq:recurrence}. Its clean derivative
is subtracted in the same way. This retains the native mixture geometry
instead of treating a normalized mixture as an unnormalized expert sum.

\subsection{Information and computation}
The predictor has white-box access to model weights, clean activations,
candidate expert outputs, and derivatives. Each prompt incurs compilation;
each queried intervention also incurs its native HVP1 calculation. Only the
structural terms are coefficient-only queries after compilation.
The full method therefore has a different cost from the structural recurrence
alone. Table~\ref{tab:costs} reports the measured components separately.
Exact edited outputs are used to evaluate predictions, not to fit or construct
Equation~\eqref{eq:ours}.

\section{Evaluation design}\label{sec:evaluation}
\subsection{Models, prompts, and intervention menus}
We evaluate the text pathways of Qwen3.6-35B-A3B and
Gemma4-26B-A4B-it \citep{qwen2026,gemma2026}. Both select eight routed
experts per token; their expert pools contain 256 and 128 experts,
respectively. The main experiments use FP32 weights and activations with
TF32 disabled; additional execution details are in Appendix~\ref{app:numerical}.
Model revisions, input identities, and scientific source hashes are pinned
in the accompanying evidence records.

The question collections contain 56 MMLU-Pro prompts, 52 SuperGPQA prompts,
and 50 text-only multiple-choice prompts from Humanity's Last Exam (HLE)
\citep{wang2024mmlu,du2025supergpqa,phan2025hle}.
The same 158 question stems are evaluated on both models; model copies do
not double the number of independent questions. MMLU-Pro contributes four
questions from each of 14 subjects. SuperGPQA contributes four from each of
13 broad disciplines. HLE contributes six or seven from each of eight broad
categories. Original options are retained, including questions with more
than four options.

MMLU-Pro uses the existing frozen 56-question cohort. The SuperGPQA and HLE
cohorts were selected by input-only hash ordering, subject coverage, and
eligibility checks, without model-performance filtering. Prompts fit within
1,024 tokens on the pinned tokenizers, require no images, and have single-token
option labels with checked prompt-boundary tokenization. Lexical duplicate
screening covers prior project inputs and the two added collections.
These subsets are evaluation inputs for intervention prediction, not official
full-benchmark accuracy evaluations. No web search or answer generation is
required to collect the next-token readout.

The source layers, in zero-based indexing, are $(27,31,35)$ for Qwen and
$(17,21,25)$ for Gemma. Source menus use clean router scores only. At each
source, we replace the weakest, middle-ranked, or strongest active expert
(ranks 8, 4, and 1) with each of the 16 highest-scoring inactive experts.
Together with the clean assignment, this gives 49 menu entries per source.
A joint condition activates two or three sources. The fixed sampling rule
covers weak, middle, strong, and mixed-strength replacements.

\subsection{Two complementary experimental panels}
\paragraph{Cross-dataset prediction.}
The main comparison uses the same 13 prespecified interventions per question
for every method: nine single replacements and four joint replacements.
This is the subset on which all five HVP derivative positions were collected
in the original protocol. The larger native panel contains the clean
condition, all 144 singles, and 32 joint conditions. We compare HVP5 only
where its full declared prediction is available, avoiding a comparison of
different candidate populations.

\paragraph{Joint ranking and native mechanism control.}
We select 24 questions from the existing 158-question collection, eight per
dataset, by input-only category and question hashes. For both models we
evaluate all 32 sampled joint combinations: 16 pairs and 16 triples.
This extends the HVP5 comparison to the complete joint pool on these inputs.
We freshly execute the same conditions twice: with natural downstream routing,
and with the nonsource downstream expert sets fixed to their clean values.
The source assignments are identical across these modes. The selected sets
are fixed in the second mode, while gate weights and the rest of the native
computation continue to respond to the intervened activations.

The complete planned panel consists of 24 questions per model, with 177
native conditions per routing mode and question. Predictions for both modes
were sealed before collecting the edited endpoints in this paired study.
The questions are reused from the larger cohort, so this panel adds a
mechanism control and a broader matched ranking comparison, rather than new
independent question stems. One separate engineering input per model is
excluded from the formal aggregates.

\subsection{Comparators and information access}
We implement the first-order attribution-patching principle, HVP1, and
multistep HVP5 in the same source-interpolation coordinates
(Equations~\eqref{eq:hvp}--\eqref{eq:multistep}).
These are adaptations of published formulas to expert replacements, rather
than reruns of the original activation-patching benchmarks. The AtP baseline
does not implement the attention Q/K and GradDrop changes in AtP$^*$.
Five-point left-Riemann integrated gradients (IG5) uses the same path
gradients as HVP5, without Hessian terms \citep{sundararajan2017ig}.

The structural comparisons are the full response recurrence,
its fixed-membership variant, and the complete predictor in
Equation~\eqref{eq:ours}. A zero-change predictor provides an effect-scale
reference. On the joint-only panel, the measured-single-sum control adds
the native effects of an intervention's constituent single replacements.
It receives actual edited single-intervention outcomes and is therefore
reported with its own information access and acquisition cost.

To test whether a denser derivative grid resolves the natural-routing error,
we also evaluate HVP20 at $0,0.05,\ldots,0.95$ on eight prespecified joint
conditions per question. HVP5, HVP20, and our predictor are compared on that
same eight-condition subset. HVP20 is not a 32-candidate ranking result.

\subsection{Metrics, uncertainty, and validation}
For each question, MAE averages absolute errors over its declared candidate
set; reported MAE then averages equally over questions. Uncertainty is shown
as a paired mean $\pm$ one standard error (SE), with the question as the
sampling unit. For question-level paired differences $d_i$, we compute
\begin{equation}
 \mathrm{SE}(\bar d)=\frac{\sqrt{\sum_h n_h s_h^2}}{N},
 \label{eq:se}
\end{equation}
where $n_h$ is the number of questions in stratum $h$, $s_h^2$ is the unbiased
sample variance of paired differences within that stratum, and $N=\sum_h n_h$.
Strata are the original subject categories for each cross-dataset comparison
and the three datasets for the joint-intervention study. This retains the
original stratum counts and pairing. Interventions within a question and
copies of a question across models are not independent replicates.
SEs are computed analytically from the recorded question-level results.

For ranking, define importance $v_i=|\Delta_x(I_i)|$. Recall@5 is the
fraction of the true five largest effects recovered by the five largest
predicted absolute effects. Importance-mass capture is the true effect mass
of the predicted top five divided by the maximum mass attainable by five
conditions. Ties use independent uniform tie breaking in forecast and truth
selection; metrics are its exact expectation. The oracle tie ceiling is
reported by the audit and equals one throughout the joint-only panel.
Uniform selection has expected recall $5/32=15.625\%$ in this 32-candidate
task; this number is not an answer-flip probability.

Validation binds inputs, candidate identities, source code, coefficients, and
forecasts by hashes. The native checks verify source assignments, the
fixed-downstream membership sets, clean-state restoration, and equality of
the interpolation endpoint with the directly assigned native execution over
the full output vocabulary. An independent reconstruction verifies the final
metrics. All reported values come from the completed, audited experiments.

\section{Prediction improves across datasets}\label{sec:results}
Table~\ref{tab:main} reports the main comparison. The complete predictor
reduces Qwen's error relative to HVP5 by 60.0\% on MMLU-Pro, 63.0\% on
SuperGPQA, and 61.7\% on HLE. Gemma's corresponding reductions are 47.4\%,
43.0\%, and 22.4\%. The relative reductions compare the dataset-level mean
errors, with identical questions and interventions for the two methods.

\begin{table*}[t]
\centering
\begingroup
\footnotesize
\renewcommand{\arraystretch}{1.10}
\setlength{\tabcolsep}{3pt}
\begin{tabular*}{\linewidth}{@{\extracolsep{\fill}}lrrrrrr@{}}
\toprule
 & \multicolumn{3}{c}{\textbf{Qwen3.6}} & \multicolumn{3}{c}{\textbf{Gemma4}} \\
\cmidrule(lr){2-4}\cmidrule(l){5-7}
Predictor & MMLU-Pro & SuperGPQA & HLE & MMLU-Pro & SuperGPQA & HLE \\
\midrule
Zero change & 0.1223 & 0.0990 & 0.0768 & 0.3001 & 0.2264 & 0.2757 \\
AtP & 0.0676 & 0.0557 & 0.0357 & 0.1478 & 0.0975 & 0.1031 \\
IG5 & 0.0629 & 0.0506 & 0.0328 & 0.1162 & 0.0743 & 0.0687 \\
HVP1 & 0.0629 & 0.0521 & 0.0334 & 0.1209 & 0.0790 & 0.0836 \\
HVP5 & 0.0626 & 0.0500 & 0.0327 & 0.1144 & 0.0722 & 0.0653 \\
\addlinespace[2pt]
Fixed response & 0.0677 & 0.0556 & 0.0356 & 0.1466 & 0.0971 & 0.1021 \\
Full response & 0.0317 & 0.0249 & 0.0179 & 0.0873 & 0.0626 & 0.0750 \\
\textbf{Ours} & \textbf{0.0250} & \textbf{0.0185} & \textbf{0.0125} & \textbf{0.0602} & \textbf{0.0412} & \textbf{0.0507} \\
\midrule
Reduction vs. HVP5 (\%) & 60.0 & 63.0 & 61.7 & 47.4 & 43.0 & 22.4 \\
$\Delta$MAE $\pm$ SE & {\scriptsize $0.0375 \pm 0.0084$} & {\scriptsize $0.0315 \pm 0.0055$} & {\scriptsize $0.0202 \pm 0.0029$} & {\scriptsize $0.0542 \pm 0.0111$} & {\scriptsize $0.0310 \pm 0.0053$} & {\scriptsize $0.0146 \pm 0.0094$} \\
Questions won & 51/56 & 44/52 & 49/50 & 50/56 & 45/52 & 36/50 \\
\bottomrule
\end{tabular*}
\endgroup

\caption{\textbf{Prediction and component comparisons.} MAE in logit units on the
same 13 interventions per question. $\Delta$MAE is HVP5 minus ours; positive
values favor our predictor. The last rows give the relative reduction,
paired mean $\pm$ SE, and number of questions with lower error. Component variants use the same natural-routing target.}
\label{tab:main}
\end{table*}

\begin{figure*}[t]
\centering\figinput{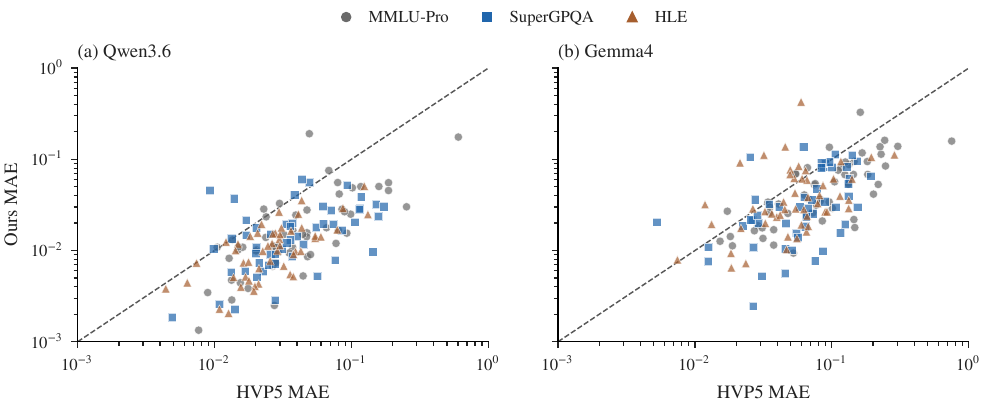}
\caption{\textbf{Paired prediction errors across all evaluation questions.}
Each point gives HVP5 MAE (horizontal) and our MAE (vertical) for one question,
averaged over its 13 matched interventions. Points below the dashed equality
line favor our predictor. Both axes are logarithmic, with identical limits
across panels. All 158 questions per model are shown; symbols and colors
identify datasets. Paired uncertainty is reported in Table~\ref{tab:main}.}
\label{fig:main}
\end{figure*}

The advantage extends across the three question collections without
retuning the predictor. Table~\ref{tab:main} gives the paired mean differences
and their SEs. All six model--dataset means favor our predictor. On Gemma--HLE,
the MAE advantage is $0.0146 \pm 0.0094$ logit units,
with the greatest relative uncertainty among the six settings. Our method
wins on 36 of that panel's 50 questions. All comparisons use equal question
weighting and the same frozen cohorts across methods.

\subsection{The structural response and curvature term contribute differently}
Table~\ref{tab:main} places the result alongside the uncorrected structural
variants and IG5. Replacing the natural gate selection by fixed membership
within the response model gives errors close to first-order prediction.
The full structural recurrence improves these values, and the complete
predictor further lowers error on all six model--dataset means. For example,
Qwen--MMLU-Pro changes from 0.0317 for the full recurrence to 0.0250 for the
complete predictor; Gemma--MMLU-Pro changes from 0.0873 to 0.0602.
This pattern is consistent with the two components addressing distinct
sources of approximation error: finite selection changes and smooth response
curvature.

The ranking experiments below test whether these improved effect estimates
change which expert combinations an analysis identifies as important.

\section{Fixing downstream expert sets isolates a major error source}\label{sec:mechanism}
The native control asks whether the derivative baseline remains inaccurate
when the intervention still changes activations and gate weights, but cannot
change nonsource expert identities. Both modes execute the pretrained model;
this is not a comparison between two variants of our response approximation.

\begin{figure*}[t]
\centering\figinput{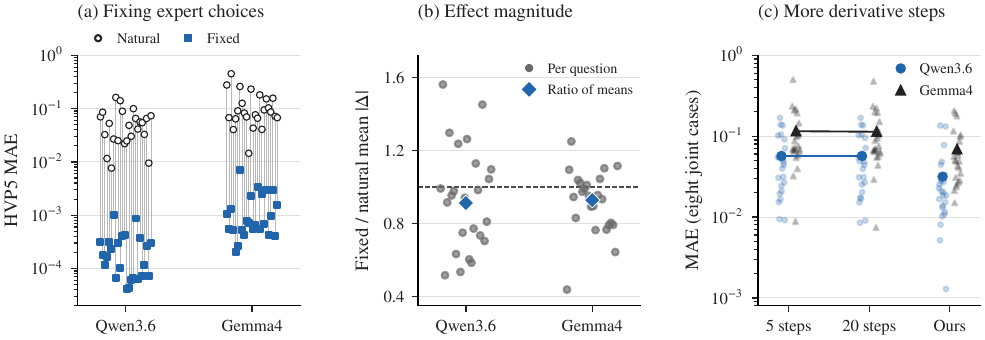}
\caption{\textbf{Question-level evidence for the effect of expert selection.}
(a) HVP5 error under natural routing (circles) and fixed later expert sets
(squares); vertical segments join the same question. (b) Fixed-to-natural
ratios of mean absolute effect: dots are per-question ratios and diamonds
are ratios of the model-level means. (c) HVP5, HVP20, and our predictor on
the same eight joint conditions: small points are questions and large markers
are means. Panels (a,b) use 32 joint conditions. Each model has 24 questions;
horizontal offsets separate observations without changing their values.}
\label{fig:clamp}
\end{figure*}

Under natural routing, HVP5 MAE is 0.055332 on Qwen and 0.121509 on Gemma.
Fixing downstream expert sets reduces these values to 0.000249 and 0.001385
(Table~\ref{tab:clamp}). The decrease occurs on every question in both models.
Mean absolute native effect sizes remain 0.0983 and 0.3718, respectively,
which are 91.2\% and 92.8\% of their natural-mode values. The decrease in
prediction error therefore greatly exceeds the modest decrease in average
intervention magnitude. Per-question normalized errors show the same pattern.

\begin{table*}[t]
\centering
\begingroup
\footnotesize
\renewcommand{\arraystretch}{1.10}
\setlength{\tabcolsep}{3pt}
\begin{tabular*}{\linewidth}{@{\extracolsep{\fill}}lrrrr@{}}
\toprule
 & \multicolumn{2}{c}{\textbf{Qwen3.6}} & \multicolumn{2}{c}{\textbf{Gemma4}} \\
\cmidrule(lr){2-3}\cmidrule(l){4-5}
Measure & Natural & Fixed & Natural & Fixed \\
\midrule
HVP5 MAE & 0.055332 & 0.000249 & 0.121509 & 0.001385 \\
Mean $|\Delta|$ & 0.1078 & 0.0983 & 0.4007 & 0.3718 \\
Normalized MAE & 0.5313 & 0.0026 & 0.3389 & 0.0034 \\
Recall@5 (\%) & 63.3 & 100.0 & 78.3 & 98.3 \\
\midrule
\multicolumn{5}{l}{\emph{Model-level routing summaries}} \\
Changed expert sets (\%) & \multicolumn{2}{c}{91.5} & \multicolumn{2}{c}{83.5} \\
Changed layers & \multicolumn{2}{c}{3.18} & \multicolumn{2}{c}{2.27} \\
Mean $|\Delta_{\rm nat}-\Delta_{\rm fix}|$ & \multicolumn{2}{c}{0.0583} & \multicolumn{2}{c}{0.1233} \\
Mean effect retained (\%) & \multicolumn{2}{c}{91.2} & \multicolumn{2}{c}{92.8} \\
\bottomrule
\end{tabular*}
\endgroup

\caption{\textbf{Native routing controls and observed routing changes.}
Natural and fixed executions use the same source assignments. Normalized
MAE averages each question's error divided by its mean absolute effect.
The lower block summarizes each model: changed sets and layers refer to
natural routing; retained effect is the ratio of fixed to natural mean
absolute effects, not a mediated-effect fraction. Recall is in percent.}
\label{tab:clamp}
\end{table*}

Changing the downstream routing rule also changes the target effect.
Consequently, the relative drop in MAE is a controlled comparison of two
executions, not an estimate that a particular percentage of the original
causal effect is mediated by routing. The paired result supports a narrower
and directly useful conclusion: HVP5 accurately predicts substantial
interventions when the nonsource expert sets are fixed, and loses much of
that accuracy when those sets can change.

\subsection{More derivative steps do not close the natural-routing gap}
On the same eight prespecified joint conditions, Qwen's MAE is 0.05692 for
HVP5 and 0.05691 for HVP20. Gemma's values are 0.11515 and 0.11344.
Our predictor has errors 0.03156 and 0.06907, respectively
(Figure~\ref{fig:clamp}c). The paired HVP20-minus-ours MAE advantages are
$0.0254 \pm 0.0106$ for Qwen and $0.0444 \pm 0.0133$ for Gemma
(mean $\pm$ SE).
Increasing the sampling density fourfold thus leaves a substantial gap in
this control. Together with the native fixed-selection experiment, this is
consistent with a finite routing response that additional smooth derivative
samples do not recover.

\subsection{Routing changes are frequent, but are not themselves effect scores}
At least one nonsource downstream expert set changes in 91.5\% of Qwen's
joint conditions and 83.5\% of Gemma's. The mean numbers of changed layers
are 3.18 and 2.27. Table~\ref{tab:clamp} also reports the mean absolute
difference between the natural and fixed native effects. These quantities
characterize the change in computation without equating the number of
changed experts with behavioral importance.

When the prediction target itself fixes downstream expert sets, the two
structural terms in Equation~\eqref{eq:ours} coincide and the estimator
reduces to HVP1. That algebra is part of the method definition; the empirical
mechanism evidence comes from the native HVP5 errors and effect magnitudes
above. Indeed, HVP5 is more accurate than HVP1 in the fixed-selection control,
as expected for the remaining smooth response.

\section{Identifying influential expert combinations}\label{sec:ranking}
For each question, an analysis must select five of 32 candidate joint
interventions using only their predicted effects. We then evaluate the
selected set against the native measurements. This task tests whether the
prediction error matters for identifying influential combinations.

\begin{figure*}[t]
\centering\figinput{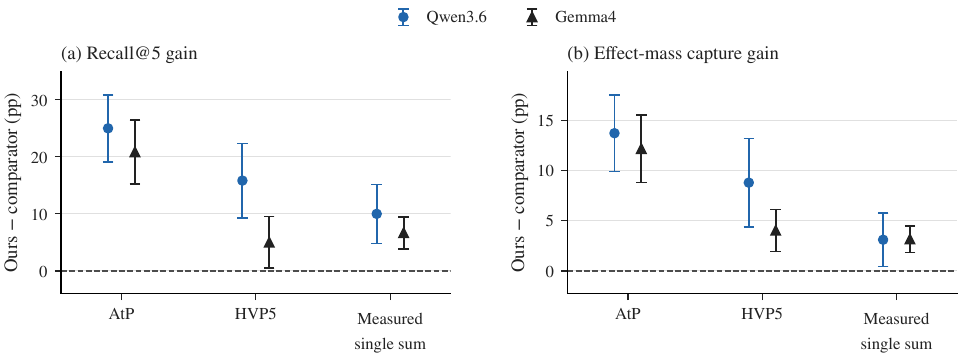}
\caption{\textbf{Paired gains in identifying influential combinations.}
Points show paired mean gains, and error bars show $\pm$ one SE.
Positive values favor our predictor; the dashed
line denotes equal performance. All comparisons use 24 questions per model
and the same 32 joint candidates. The single-sum reference uses measured
single-intervention effects. Gemma's recall gain against HVP5 is comparable to its SE.}
\label{fig:ranking}
\end{figure*}

Table~\ref{tab:joint} reports both error and ranking. Relative to HVP5,
the complete predictor reduces joint-only MAE by 45.1\% on Qwen and
41.0\% on Gemma. Recall@5 rises from 63.3\% to 79.2\% and from 78.3\%
to 83.3\%. Thus, on Qwen, the mean recovered set contains approximately
3.96 of the five true top interventions, compared with 3.17 for HVP5.
The reference is measured native effect size, rather than agreement with
another predictor's ranking.

\begin{table*}[t]
\centering
\begingroup
\footnotesize
\renewcommand{\arraystretch}{1.10}
\setlength{\tabcolsep}{3pt}
\begin{tabular*}{\linewidth}{@{\extracolsep{\fill}}lrrrrrr@{}}
\toprule
 & \multicolumn{3}{c}{\textbf{Qwen3.6}} & \multicolumn{3}{c}{\textbf{Gemma4}} \\
\cmidrule(lr){2-4}\cmidrule(l){5-7}
Predictor & MAE & Recall@5 & Mass & MAE & Recall@5 & Mass \\
\midrule
\multicolumn{7}{l}{\emph{Predictors without measured single effects}} \\
AtP & 0.0687 & 54.2 & 79.3 & 0.1741 & 62.5 & 85.3 \\
HVP1 & 0.0579 & 65.8 & 85.2 & 0.1260 & 71.7 & 90.1 \\
HVP5 & 0.0553 & 63.3 & 84.3 & 0.1215 & 78.3 & 93.5 \\
Full response & 0.0396 & 71.7 & 89.1 & 0.1256 & 65.0 & 87.9 \\
\textbf{Ours} & \textbf{0.0304} & \textbf{79.2} & \textbf{93.1} & \textbf{0.0717} & \textbf{83.3} & \textbf{97.5} \\
\addlinespace[2pt]
Gain vs. HVP5 & $0.0250 \pm 0.0082$ & $15.8 \pm 6.5$ & $8.8 \pm 4.4$ & $0.0498 \pm 0.0130$ & $5.0 \pm 4.5$ & $4.0 \pm 2.1$ \\
\midrule
\multicolumn{7}{l}{\emph{Reference using measured single effects}} \\
Single sum & 0.0483 & 69.2 & 90.0 & 0.0914 & 76.7 & 94.3 \\
Gain vs. single sum & $0.0179 \pm 0.0074$ & $10.0 \pm 5.2$ & $3.1 \pm 2.7$ & $0.0196 \pm 0.0107$ & $6.7 \pm 2.8$ & $3.2 \pm 1.3$ \\
\bottomrule
\end{tabular*}
\endgroup

\caption{\textbf{Joint-intervention accuracy, ranking, and paired uncertainty.}
Recall@5 and effect-mass capture are percentages; MAE is in logit units.
Each gain row names its comparator and reports a paired mean $\pm$ SE.
For MAE, gain is comparator minus ours; for ranking, it is ours minus
comparator in percentage points. Differences use values before rounding.
The separate single-sum reference adds measured constituent effects.}
\label{tab:joint}
\end{table*}

Qwen's paired recall improvement is $15.8 \pm 6.5$ percentage points
(mean $\pm$ SE). Gemma's improvement is $5.0 \pm 4.5$ points,
so its mean gain is comparable to its uncertainty. Effect-mass capture
increases from 84.3\% to 93.1\% on Qwen and from 93.5\% to 97.5\% on Gemma;
the paired gains and SEs are in Table~\ref{tab:joint}. This metric accounts
for cases where the selected set misses a top-five identity but recovers
a nearly equal effect.

The measured-single-sum control obtains recall 69.2\% on Qwen and 76.7\%
on Gemma, with errors 0.0483 and 0.0914. The complete predictor improves
both model means without receiving the measured single-intervention effects.
This comparison is useful because a single intervention can already include
its own downstream route changes. Its additive composition still need not
recover how several source replacements interact.

\subsection{Ranking on the larger question collection}
We also retrospectively evaluate top-five recall on the original 13-condition
matched panel (Table~\ref{tab:ranking13}, Appendix~\ref{app:ranking}). The correction improves the mean
recall in every model--dataset group. This analysis uses all 158 questions per
model but mixes single and joint interventions and has a smaller candidate
pool. It is supporting evidence, with separate sampling and interpretation
from the joint-only mechanism study.

\subsection{Accuracy and execution cost}
Table~\ref{tab:costs} records mean timing components for the joint-only panel
on the A100 80GB execution. The correction uses the HVP1 probe costs plus
prompt compilation and structural queries. HVP5 uses five native derivative
positions per candidate. Direct edited forward execution is much cheaper
than derivative computation in the current cached-suffix implementation.
The ranking results therefore establish improved identification of important
interventions from predictions; they do not establish an end-to-end search
speedup over executing all 32 candidates.

\section{Relation to prior work}\label{sec:related}
\paragraph{Sparse expert computation.}
Sparsely gated MoE layers select a small subset of expert functions for each
input \citep{shazeer2017moe}. GShard and Switch Transformers develop this
conditional-computation approach at larger scales
\citep{lepikhin2020gshard,fedus2022switch}. We study the predictive consequence
of this selection mechanism: an expert intervention can change which
functions later layers execute.

\paragraph{Causal evidence and circuit discovery.}
Causal tracing and activation patching connect language-model behavior to
internal components, including factual recall and indirect object
identification \citep{meng2022rome,wang2022ioi}. ACDC automates circuit
selection through interventions \citep{conmy2023acdc}, while edge attribution
patching uses gradient estimates to score candidate edges
\citep{syed2024eap}. EAP-IG evaluates integrated-gradient estimates through
circuit faithfulness \citep{hanna2024faithfulness}. These studies motivate
checking what a prediction-based selection procedure omits. Our evaluation
measures recovery of influential expert combinations in a specified candidate
pool.

\paragraph{Intervention design and readouts.}
Patching results depend on choices such as the intervention and output metric
\citep{zhang2024best,heimersheim2024patching}. We fix the source assignments,
prompt position, and output-token contrast before measuring intervention
effects. This makes the natural-versus-fixed routing comparison explicit and
keeps the prediction target identical across methods within each mode.

\paragraph{Gradient estimates and interactions.}
Attribution patching approximates activation-patching effects with gradients.
AtP$^*$ addresses attention saturation and cancellation through Q/K corrections
and GradDrop \citep{kramar2024atp}. Our AtP comparator uses its first-order
principle. Integrated gradients attributes endpoint differences through path
derivatives \citep{sundararajan2017ig}; Integrated Hessians extends this
approach to pairwise feature interactions \citep{janizek2021interactions}.
HVP and multistep HVP estimate response curvature \citep{zhang2026hvp}.
We adapt the latter formulas to a common expert-interpolation path that
includes mixed-source Hessian terms. The structural contribution estimates
the additional response from changes in downstream expert identities.

\paragraph{Models of component effects.}
COAR predicts component-intervention effects and connects them to model
editing \citep{shah2024coar}. Circuit tracing builds attribution graphs in
feature-based replacement models and tests graph hypotheses through
interventions \citep{ameisen2025tracing}. Our compiled representation uses
the pretrained model's expert outputs and native gates. It predicts the
score effect of a specified expert replacement without fitting its measured
outcome.

\paragraph{Routing and the meaning of an attribution.}
\citet{nguyen2026contract} study how the score, conditioning choices, and
generation process determine what an attribution explains. Their MoE
experiment shows that recomputed routing can leave an integrated-gradients
residual that fixing routing reduces. We retain natural downstream routing
as the prediction target and estimate the effect of expert reselection.
Our native control, effect predictions, and ranking comparisons test that
estimate at the declared expert-intervention sites.

\paragraph{Intervening on routing.}
Counterfactual routing has also been used to examine or change MoE behavior
\citep{hu2026dormant,yoon2026misrouted}. We leave model parameters fixed and
predict the effect of an assigned expert replacement. Actual edited-model
responses provide the reference for evaluating these predictions and their
use in identifying influential combinations.

\section{Discussion and scope}\label{sec:discussion}
The experiments support an explicit correction for the response introduced by
downstream expert reselection. Three observations align: the natural-routing
prediction improves when that correction is included, native fixed-membership
executions become easy for HVP5 to predict, and denser HVP sampling leaves
the natural-routing error largely intact. The ranking results show that the
accuracy improvement can change which expert combinations are identified as
important. Together these observations connect the predictor's structure to
both a mechanism and an analysis outcome.

\paragraph{Local approximation and computational cost.}
Clean expert bases and clean transport approximate the content response away
from the original trajectory. Strong content changes can therefore remain
difficult even when routing is modeled. The full predictor also requires
per-candidate second derivatives. Its present computation costs motivate
reducing compilation overhead and amortizing curvature calculations before
claiming a practical speed advantage over exact native execution.

\paragraph{What the evaluation establishes.}
The main evidence covers two MoE families, three question subsets, and finite
replacements at three source layers of the final prompt position.
The 24-question mechanism panel reuses the main cohort. It is a controlled
within-input comparison, not an independent replication on new text.
The readout is a fixed next-token logit contrast; a different readout,
intervention family, source location, or generation process can change the
prediction problem. In particular, the present ranking experiment measures
identification of large score effects, rather than a guarantee of answer
flips, answer correction, or long-form behavioral control.

\section{Additional ranking results}\label{app:ranking}
Table~\ref{tab:ranking13} gives the retrospective ranking comparison on all
158 shared question stems. Each question has 13 candidates: nine single
replacements and four joint replacements. This panel therefore differs from
the 32-candidate joint-only study in Section~\ref{sec:short-results}. We report both the
identities of the true top five that are recovered and the true effect
magnitude captured by the selected set. Gains are paired means $\pm$ SE,
computed from question-level differences within the original subject strata.
The main prediction means, component comparisons, and paired MAE gains are
consolidated in Table~\ref{tab:main}.

\begin{table*}[t]
\centering
\begingroup
\footnotesize
\renewcommand{\arraystretch}{1.10}
\setlength{\tabcolsep}{3pt}
\begin{tabular*}{\linewidth}{@{\extracolsep{\fill}}lrrrrrr@{}}
\toprule
 & \multicolumn{3}{c}{Recall@5 (\%)} & \multicolumn{3}{c}{Effect mass captured (\%)} \\
\cmidrule(lr){2-4}\cmidrule(l){5-7}
Dataset & HVP5 & Ours & Ours $-$ HVP5 & HVP5 & Ours & Ours $-$ HVP5 \\
\midrule
\multicolumn{7}{l}{\textbf{Qwen3.6}} \\
MMLU-Pro & 73.9 & 90.4 & $16.4 \pm 2.2$ & 90.3 & 97.4 & $7.1 \pm 1.4$ \\
SuperGPQA & 74.2 & 87.3 & $13.1 \pm 2.7$ & 88.5 & 97.1 & $8.7 \pm 1.8$ \\
HLE & 72.0 & 88.8 & $16.8 \pm 2.5$ & 87.9 & 98.1 & $10.2 \pm 1.4$ \\
\addlinespace[3pt]
\multicolumn{7}{l}{\textbf{Gemma4}} \\
MMLU-Pro & 75.0 & 86.8 & $11.8 \pm 1.9$ & 90.3 & 96.8 & $6.5 \pm 1.2$ \\
SuperGPQA & 75.8 & 87.3 & $11.5 \pm 2.6$ & 90.5 & 97.5 & $7.1 \pm 1.5$ \\
HLE & 86.0 & 90.0 & $4.0 \pm 2.1$ & 95.9 & 98.9 & $3.0 \pm 1.1$ \\
\bottomrule
\end{tabular*}
\endgroup

\caption{\textbf{Ranking on the complete question collection.} Recall and
captured effect mass on the 13-candidate panel. Gains are ours minus HVP5,
in percentage points, reported as mean $\pm$ SE. Each model uses the same 56 MMLU-Pro,
52 SuperGPQA, and 50 HLE questions; the methods use identical candidates.}
\label{tab:ranking13}
\end{table*}

\section{Computation and reproduction details}\label{app:reproduction}
Table~\ref{tab:costs} separates per-prompt compilation from the derivative
and native-execution costs for 32 joint candidates. The complete predictor
requires HVP1 queries in addition to compilation and structural queries.
These recorded components exclude model loading, prompt preparation, I/O,
and auditing; they are not end-to-end runtime measurements.

\begin{table}[t]
\centering
\begingroup
\footnotesize
\renewcommand{\arraystretch}{1.10}
\setlength{\tabcolsep}{3pt}
\begin{tabular}{@{}lrrrr@{}}
\toprule
Model & Compile & HVP1 & HVP5 & Forward \\
\midrule
Qwen3.6 & 67.3 & 44.3 & 219.7 & 2.33 \\
Gemma4 & 68.9 & 10.7 & 52.4 & 2.10 \\
\bottomrule
\end{tabular}
\endgroup

\caption{\textbf{Recorded computation costs in seconds.} Compile is a per-prompt
cost. HVP1, HVP5, and direct Forward cover all 32 joint candidates. Means are
from the A100 80GB study.}
\label{tab:costs}
\end{table}

\paragraph{Frozen intervention design.}
Each of three source layers has 49 menu entries: the clean assignment and
48 single-expert replacements. The replaced clean expert has rank 8, 4,
or 1 by clean router score, and its replacement is one of the 16
highest-scoring inactive experts. The 144 single conditions and the
32 sampled joint conditions form a 177-condition native panel with the
clean baseline. Every designated source is explicitly assigned in all
conditions, including the sources whose menu entry is zero.

\paragraph{The two evaluation panels.}
The common 13-condition prediction panel contains nine singles and four
joint interventions per question. Its question collections contain 56,
52, and 50 questions from MMLU-Pro, SuperGPQA, and HLE. The paired native
mechanism study selects eight existing questions from each collection by
input-only category and question hashes, giving 24 shared stems and
48 model--question contexts. The 32 joint conditions contain 16 pairs
and 16 triples. HVP20 is available for eight predeclared joint conditions
per question, so its comparison uses that same subset for every method.

\paragraph{Statistical and execution units.}
Errors first average within question and then equally across questions.
Paired SEs retain the subject strata in the cross-dataset study and the
dataset strata in the mechanism study, as defined in Equation~\eqref{eq:se}.
Models are reported separately. Native endpoints,
assigned expert sets, and the preservation of clean execution are checked
against the frozen intervention definitions. These tables reuse the completed
reports without additional experimental MoE executions, fitting, or
inferential tests.

\section{Additional execution details}\label{app:numerical}
\paragraph{Numerical configuration.}
The main experiments use FP32 pretrained weights and activations, eager
native operators, and disabled TF32. Compiled transport and readout
accumulation use FP64 arithmetic; gate queries retain the original FP32
gate program.

\paragraph{Earlier measurements.}
The main comparisons use the execution configuration above. Earlier BF16 measurements on the
56-question MMLU-Pro cohort give HVP5 versus corrected errors of
0.1281 versus 0.1199 on Qwen and 0.1631 versus 0.1396 on Gemma, smaller
relative advantages than the FP32 results. Those precision-specific runs
are retained in the evidence; they do not establish that the FP32 advantage
transfers unchanged to BF16. Finite-precision score rounding and its relation
to derivatives require separate treatment. The real-arithmetic argument in
Proposition~\ref{prop:jumps} isolates expert selection and is not a theorem
about all errors in a quantized execution.

\FloatBarrier

\end{document}